\documentclass[conference]{IEEEtran}
\usepackage{blindtext, graphicx}
\usepackage{amsmath}
\pdfoutput=1
\usepackage{doi}
\usepackage{balance}
\usepackage{caption} 
\ifCLASSINFOpdf
\else
\fi

\IEEEoverridecommandlockouts
\begin{document}
%
\title{Continuous Multimodal Emotion Recognition Approach for AVEC 2017}

\author{\IEEEauthorblockN{Narotam Singh\thanks{\footnotemark{*} \IEEEauthorrefmark{2} Third year undergraduates - Equal Contribution}\footnotemark{*},
Nittin Singh\IEEEauthorrefmark{2}, Abhinav Dhall\IEEEauthorrefmark{3}}
\IEEEauthorblockA{Department of Computer Science and Engineering,
Indian Institute of Technology Ropar\\
India\\
Email: \footnotemark{*}2015csb1065@iitrpr.ac.in,
\IEEEauthorrefmark{2}2015csb1067@iitrpr.ac.in,
\IEEEauthorrefmark{3}abhinav@iitrpr.ac.in}}


%



\maketitle

\begin{abstract}
This paper reports the analysis of audio and visual features in predicting the continuous emotion dimensions under the seventh Audio/Visual Emotion Challenge (AVEC 2017), which was done as part of a B.Tech. 2nd year internship project. For visual features we used the HOG (Histogram of Gradients) features, Fisher encodings of SIFT (Scale-Invariant Feature Transform) features based on Gaussian mixture model (GMM) and some pretrained Convolutional Neural Network layers as features; all these extracted for each video clip. For audio features we used the Bag-of-audio-words (BoAW) representation of the LLDs (low-level descriptors) generated by openXBOW provided by the organisers of the event. Then we trained fully connected neural network regression model on the dataset for all these different modalities. We applied multimodal fusion on the output models to get the Concordance correlation coefficient on Development set as well as Test set.
\end{abstract}


%
\IEEEpeerreviewmaketitle

\section{Introduction}
Emotions are one of the most important and complex aspects of human consciousness. Researchers have been trying to classify or measure human emotions, as this research can lead to a significant improvement in human-computer interaction. Many studies have defined emotions as discrete categories. In discrete emotion theory, humans are thought to have some basic emotions. These basic emotions include ‘anger’, ‘disgust’, ‘fear’, ‘happiness’, ‘sadness’ and ‘surprise’ \cite{discrete}. While some other studies represent emotions using continuous dimensional models. Dimensional models of emotion attempt to represent emotions as a particular continuous region in 2-D or 3-D continuum. Harold Schlosberg in his study defined the three dimensions of emotion: ‘pleasantness-unpleasantness’, ‘attention-rejection’, ‘level of activation’ \cite{dim1}. Dimensional  analysis of emotions suggest that each dimension/factor is related to different regions in our brain, and are processed independently. Most studies incorporate two dimensions that are ‘arousal’ and ‘valence’. ‘Valence’ as used in psychology of emotions represents ‘pleasantness-unpleasantness’ of an emotion, while ‘Arousal’ signifies the level of activation. Another dimension is ‘liking’ which points the wanting of a particular experience \cite{liking}. Recently the dimensional models are being widely used for research purposes, thanks to their practicality and ease of representing complex emotions. The basic motivation behind this challenge is the observation that a lot of human emotions can be judged by the physical gestures of human face while interacting with other people. \cite{psychology} Psychologists have shown that various facial activities like eye movements, lip movement, wrinkles around eye-corners etc. have special semantic meanings in emotional context, and conversely by detecting these facial signs one can judge the emotional state of a person. Various studies have shown that it is in fact how humans understand others emotions.\\This paper implements our work in the challenge and our main focus is on the visual and audio features. We have extracted the visual features in the form of HOG features, deep visual features and fisher vector representation of SIFT \cite{sift} features and the audio features are used as it is provided by the organizers of the challenge. Then we have trained models using Neural Networks.
\section{Related Works}
Affective Computing Analysis has been attracting many researchers since it was first introduced into modern computing by Rosalind Picard in 1995 \cite{affective}. Since then researchers have been introducing new techniques to predict human emotions. Regarding the extraction of features using audio signals and from human speech, feature sets such that LPC \cite{dim} have been used. MFCC \cite{fishk} features have also been used by some getting better results. Teams which participated in AVEC 2016 competition \cite{Povolny:2016:MER:2988257.2988268} also used physiological features like the heart rate (HR), ECG signals, the skin conductance response (SCR), the skin conductance level (SCL), etc. Apart from the visual features like HOG \cite{hog}  and SIFT \cite{sift} features, PHOG are also successful in human detection \cite{huk}. HOG \cite{hog} and LBP (Local binary patterns) features can also be used in combination like in \cite{lop}, where they complemented each other. LBP tend to reduce the noise of hog noisy edges. LSTM (Long Short Term Memory) is a type of Recurrent neural Network proposed in 1997 which was used for affect recognition in \cite{arf}. This network was also used by teams in the AVEC 2015 competition \cite{He:2015:MAD:2808196.2811641}. Talking of neural networks, deep neural networks like Deep Convolutional Neural Network in \cite{nnn}  used transductive learning approach for emotion recognition along with the hypergraphs.\\
Multimodal techniques are  nowadays and are being used widely in research areas \cite{multi}. Researchers dealing with the different types of modalities very often want to get better results by combining their results. Many such techniques exist, one of them being Markov fusion Network \cite{mfn}. MFN combines classifier outputs of different modalities  with respect to its temporal relationship.
Researchers also created apps and softwares for sentiment prediction that analyses facial expressions, gestures, speech, etc. One such tool is Kairos Emotions API\footnote{https://www.kairos.com/emotion-analysis-api} which is available online.

\section{Dataset}
This challenge is based on a novel database of human-human interactions recorded `in-the-wild': Automatic Sentiment Analysis in the Wild (SEWA\footnote{https://sewaproject.eu/}) \cite{schuller2016multimodal} data set. The dataset contains the video data, audio data and manual transcriptions of the speech of the corresponding subjects for training, development and testing with emotion dimensions values : arousal, valence and liking. Among the audio data are the audio files of the subjects, 23 LLDs from the eGeMAPSv0.1a feature set, 88 acoustic features from the eGeMAPSv0.1a feature set and  Bag-of-audio-words representation (BoAW) of the LLDs for all the subjects. Among the video data are the video files of the subjects, video features (pixel coordinates of facial landmark points and head orientation) : normalised and unnormalised, and Bag-of-video-words representation of the normalised video features.
\section{Features}
\subsection{Audio Features}
The BoAW representation (generated by openXBOW \cite{xbow}) of 23 LLDs from eGeMAPSv0.1a \cite{eg} feature set extracted using openSmile \cite{opensmile} are provided in the sub challenge dataset. We used BoAW with block size 6 seconds as it is for the training process using neural network.
\subsection{Visual Features}
Before extracting any visual feature, we detected the human faces from the frames of the videos and applied the affine transformation using the facial landmark coordinates provided in the dataset to get the vertical faces. Faces are detected using the Viola-Jones face detection \cite{Viola01robustreal-time} algorithm and then cropped. All the visual features further discussed are extracted from these affine-warped faces. Most of the visual features are trained after normalisation process except in the cases where it increased the error. Images are resized and converted to gray scale as per need depending on the model requirements.
\subsubsection{Histogram of Gradients (HOG) features}
Faces are resized to the default window size of HOG \cite{hog} (64x128) so as to get a single feature from the frames.
Then trained on the neural network after normalisation.
\subsubsection{SIFT-Fisher Vector representation with GMM}
Bag of words model is a quite successful approach of simplifying the representation of text/documents, used in Natural Language Processing. In the context of computer vision it has proved to be a very effective representation of an image \cite{david}.
We have used the BOW representation with GMM. GMM is a generative and rather complex clustering model as compared to simpler models such as k-means. We used GMM model for final cluster formation as it takes into account the distance of a data point from the centroid and also the weight of a particular cluster, which is otherwise ignored in simple k-means.
We first calculated the SIFT \cite{sift} descriptors for each frame and only the descriptors having highest response value were chosen (top 50 descriptors). Then we clustered our entire data into 32 clusters\footnote{The cluster number value 32 was chosen empirically.} using K-means algorithm.  After dividing the data into 32 clusters, we calculated mean, covariance matrix and weights for each cluster. Weights are calculated as the fraction of data points allotted to that cluster. After that these values are used as the initial values of means, covariance matrices and weights in the EM (Expectation-Maximization) algorithm for the generation of the GMM. \\
\begin{figure}[h]
\caption{GMM for simple 2D data visualization.}
\begin{center}
\includegraphics[scale=0.1]{{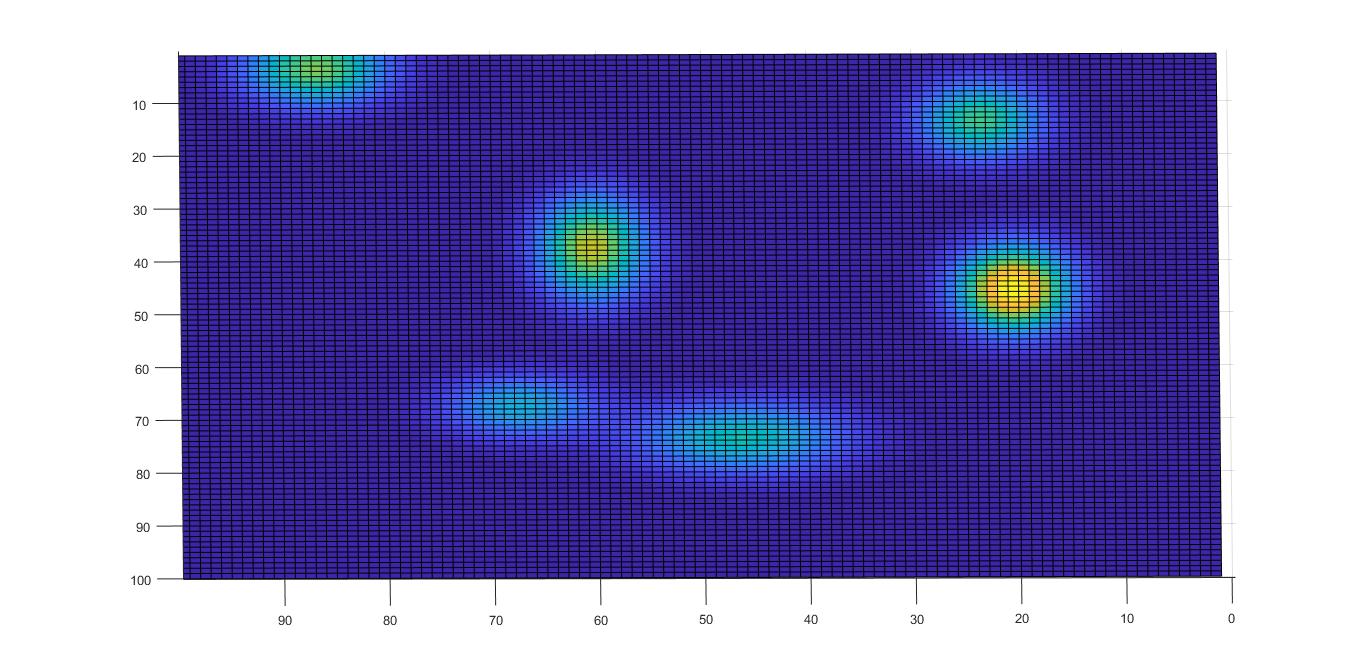}}
\includegraphics[scale=0.1]{{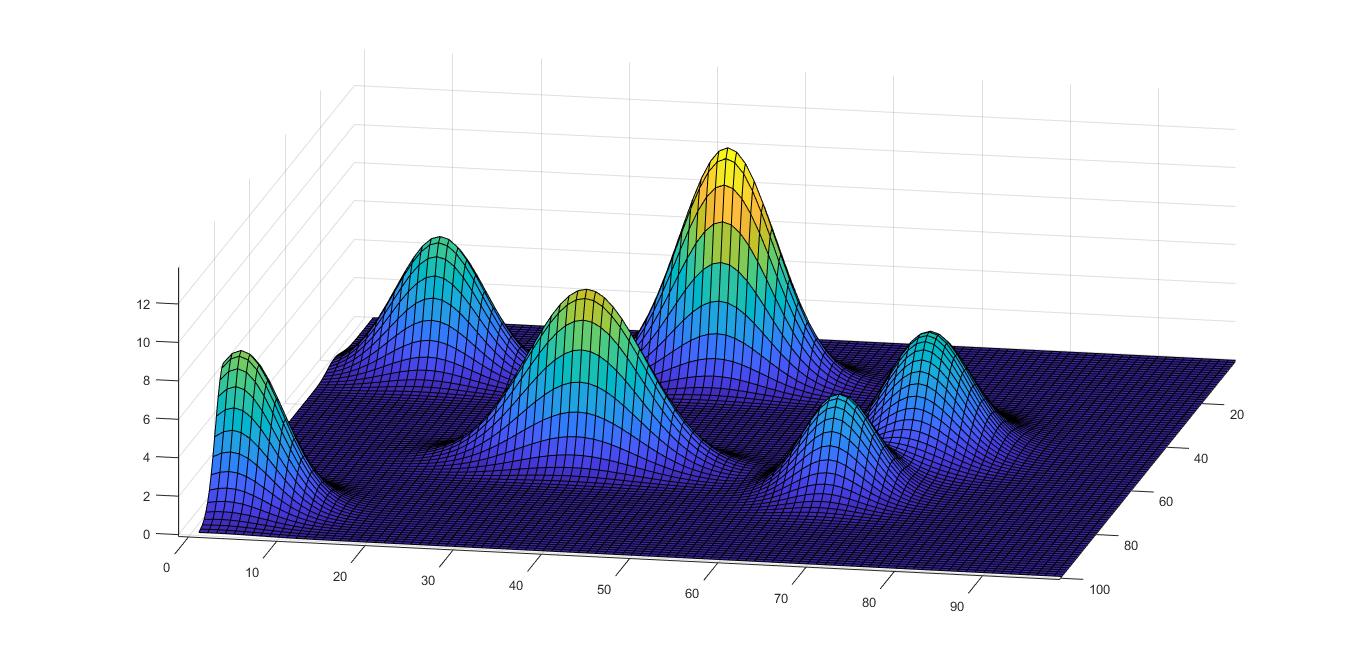}}
\end{center}
\end{figure}
\paragraph{Fisher Vector Extraction}
Following the fisher vector approach in \cite{7344580}, we took the previously calculated SIFT features of each frame and got the Fisher vector encodings \cite{fish1} of these sift features based on the pre-calculated GMM model. 
The resultant fisher vector has dimensions 128x32x2. Dimension reduction technique Principal Component Analysis (PCA) is used to convert the fisher encoding into final 4508 dimensions (99 percent variance). Thus a final fisher vector is obtained for each frame of each video.
\subsubsection{Deep Visual Features}

An output of a particular layer of pretrained models : VGG-Face \cite{vgg} and ResNet-50-dag \cite{He2015} are used as features. These models are deep convolutional neural networks obtained from MatConvNet toolbox \cite{vedaldi15matconvnet}. Output of 35\textsuperscript{th} layer of VGG-Face and output of 174\textsuperscript{th} layer of ResNet-50-dag \cite{He2015} are used as features of the frames of each video clip. These are activation layer (ReLU) and a fully connected layer in their respective models.

\section{Training}
Three different models for arousal,valence and liking are trained for different modalities. We used Neural Networks for this regression task instead of LibSVM because it was very time consuming for such a large dataset. More or less the number of layers used in the network is the same in all the models. ReLU (rectified linear unit) activation function is used between layers except the last layer where \(tanh\) function is used to get the final output. We observed that 4-5 layers were sufficient and increasing the layers more than this was computationally expensive as well as it did not improve the results.
\section{Results}
For each emotion dimension, the models trained on the Train Set, are used for the prediction on the Development (Dev) Set. Then CCC is calculated for each model. Here \( \rho\) is the correlation coefficient between two variables \(x\) and \(y\), and \(\sigma_x^2 , \sigma_y^2\) is the variance of \(x\) and \(y\) respectively while \(\mu_x , \mu_y\) are respective means. Here CCC is the Concordance Correlation Coefficient which is calculated as:\vspace{3mm}\[CCC = \frac{2\rho\sigma_x\sigma_y}{\sigma_x^2 + \sigma_y^2 + (\mu_x-\mu_y)^2}\]

\begin{table}[h!]
\label{table:1}
\captionsetup{format=hang}
\caption{AVEC Baseline \cite{ringeval2017avec} : Concordance Correlation Coefficient Table for Development (Dev) Set and Test Set.}
\centering
 \begin{tabular}{|c|| c| c| c|} 
 \hline
 Model & Arousal & Valence & Liking \\ [0.5ex] 
 \hline\hline
 Dev-Audio & 0.344 & 0.351 &0.081     \\ 
 \hline
 Dev-Video & 0.466 & 0.400 & 0.155 \\
 \hline
 Dev-Text & 0.373  & 0.390 & 0.314 \\
 \hline
 Dev-Multimodal & 0.525 & 0.507 & 0.235 \\
 \hline
  Test-Audio & 0.225 & 0.244 & -0.020 \\
 \hline
  Test-Video & 0.308 & 0.455 & 0.002 \\
 \hline
  Test-Text & 0.375 & 0.425 & 0.246 \\
 \hline
  Test-Multimodal & 0.306 & 0.466 & 0.048 \\
 \hline
\end{tabular}

\end{table}
\begin{table}[h!]
\label{table:2}
\captionsetup{format=hang}
\caption{Obtained: Concordance Correlation Coefficient Table for Development (Dev) Set.}
\centering
 \begin{tabular}{|c|| c| c| c|} 
 \hline
 Model & Arousal & Valence & Liking \\ [0.5ex] 
 \hline\hline
 HOG & 0.289 & 0.310 &0.056     \\ 
 \hline
 VGG FACE & 0.277 & 0.336 & 0.021 \\
 \hline
 RESNET & 0.165  & 0.274 & 0.010 \\
 \hline
 SIFT-fisher & 0.075 & 0.123 & -0.021 \\
 \hline
 AUDIO & 0.212 & 0.218 & -0.089\\
 \hline

\end{tabular}

\end{table}
\section{Multimodal Fusion}
Now we can perfom multimodal fusion on these features to get final fused results.
\begin{figure}[h]

\includegraphics[scale=0.7]{{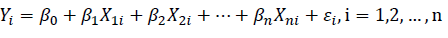}}
\end{figure}
\begin{table}[h!]
\label{table:3}
\captionsetup{format=hang}
\caption{Concordance Correlation Coefficient Table of the Fused results.}
\centering
 \begin{tabular}{|c|| c| c| c|} 
 \hline
 Fused Features & Arousal & Valence & Liking \\ [0.5ex] 
 \hline\hline
  Dev-Multimodal & 0.294 & 0.346 & 0.013\\
  \hline
  Test-Multimodal & 0.276 & 0.365 & 0.00\\
 
 \hline
\end{tabular}
\end{table}
\begin{table}[h!]
\captionsetup{format=hang}
\caption{Linear Correlation Coefficient for Test Data.}
\centering
 \begin{tabular}{|c|| c| c| c|} 
 \hline
 Feature & Arousal & Valence & Liking \\ [0.5ex] 
 \hline\hline
  Test-Multimodal & 0.361 & 0.437 & -0.001\\
 
 \hline
\end{tabular}
\end{table}
\section{Conclusion}
This report summarizes our second year internship project which proved very informative as beginners and this was our first time handling such large datasets. Although the field of emotion recognition is not in very advanced state, still the rate of growth ensures a quite bright future.\\ From our project we can say that visual features such as HOG, SIFT, etc. are capable of capturing the important facial information. We observed that the HOG features and VGG-Face features performed quite well as compared to the others.We were expecting better results from the SIFT-GMM features but the results obtained are not very encouraging. However the overall results have shown that these visual features are quite dependable, for most of the cases and can be used as a way for emotion detection. Moreover, it was observed that the multi-modal fusion provides even better results by giving suitable weights to the respective features. Further we have observed that the weights varied from model to model, which suggests that the success of detecting a particular emotion is dependent on the type of feature we are extracting, and simpler models can have more impact as compared to complex models. 

\section*{Acknowledgment}
We are greatly thankful to our teacher and the entire AVEC 2017 organising team, who gave us the opportunity to participate in this informative event. This project has helped us in gaining more knowledge and insight regarding the field of computer vision as well as machine learning and some basic knowledge of deep learning.




%
\vspace{1cm}
\bibliographystyle{acm}
\balance
\bibliography{bib}

\end{document}